 \documentclass[sigconf,nonacm]{acmart}

\AtBeginDocument{%
  \providecommand\BibTeX{{%
    \normalfont B\kern-0.5em{\scshape i\kern-0.25em b}\kern-0.8em\TeX}}}

\setcopyright{acmcopyright}

\acmConference[CODS-COMAD '22]{}{January 08--10, 2022}{Bangalore, India}
\acmBooktitle{CODS-COMAD '22,January 08--10, 2022, Bangalore, India}

\usepackage{url}  
\usepackage{graphicx}  
\usepackage{wrapfig}  
\usepackage{multicol}

\usepackage{xcolor}
\usepackage[linesnumbered,ruled,vlined]{algorithm2e}

\newcommand{\eat}[1]{#1}
\newcommand{\add}[1]{#1}
\newcommand{\cancel}[1]{}
\newcommand{\remove}[1]{}
\newcommand{\cancelpc}[1]{#1}
\newcommand{\system}{AITEST}

\usepackage{url}
\usepackage{wrapfig}
\usepackage{amsmath}
\usepackage{graphics}
\usepackage{graphicx}
\usepackage{latexsym}
\usepackage{array}
\usepackage{enumitem}
\newcolumntype{M}[1]{>{\centering\arraybackslash}m{#1}}
\usepackage{multirow}
\usepackage{url}
\usepackage{listings,color,colortbl}

\definecolor{light-gray}{gray}{0.80}

\usepackage{balance}

\begin{document}

\title{Data Synthesis for Testing  Black-Box Machine Learning Models}

\author{Diptikalyan Saha}
\affiliation{%
  \institution{IBM Research, India}
  \country{}}
\email{diptsaha@in.ibm.com}

\author{Aniya Aggarwal}
\affiliation{%
  \institution{IBM Research, India}
  \country{}}
\email{aniyaagg@in.ibm.com}

\author{Sandeep Hans}
\affiliation{%
  \institution{IBM Research, India}
  \country{}}
\email{shans001@in.ibm.com}

\begin{abstract}
The increasing usage of machine learning models raises the question of the reliability of these models. The current practice of testing with limited data is often insufficient. In this paper, we provide a framework for automated test data synthesis to test black-box ML/DL models. We address an important challenge of generating realistic user-controllable data \cancelpc{with model agnostic coverage criteria }to test a varied set of properties, essentially to increase trust in machine learning models. We experimentally demonstrate the effectiveness of our technique. 
\end{abstract}

\cancel{
}

\maketitle

\section{Introduction}
\label{sec:intro}
Artificial Intelligence systems are increasingly being used in critical applications, such as to assess a criminal defendant’s likelihood of committing a crime~\cite{recidivism}, selecting resumes for recruitment~\cite{resume}, loan processing~\cite{loan}, etc.  While AI holds the promise of delivering valuable insights and knowledge across a multitude of applications, the broad adoption of AI systems will rely heavily on the ability to trust their output. 

In order to assure  the reliability of AI systems, in this paper, we address some of shortcomings of automated testing of AI models, as described below. 
\cancel{
 A natural question arises regarding how to ensure the reliability of an AI system. Because of its simplicity/understandability, efficiency,  and usefulness for finding faults,  testing~(\cite{EXE,DART,DEEPXPLORE}) remains a universal choice for checking the reliability of a system. Most industrial practices have a dedicated phase and personnel in the software development life-cycle for testing. In this paper, we, therefore, investigate the area of AI Testing i.e. Testing of AI models (rather than using AI techniques for Testing).}
\
The data scientists use the technique of splitting the cleaned data to get train and test set in order to build and select the best model in terms of accuracy.
In industry, Model Risk Management~\cite{MRM} phase of the Data and AI lifecycle  independently tests the model on various characteristics such fairness, robustness, drift, and  business KPIs before deploying the model. Such risk management imposes the following additional requirements.

\begin{itemize}[leftmargin=*]

\item Risk management requires additional realistic test data, not used by data scientists as the available test split may not be a complete representation of the possible payload data making it insufficient.

\item Simulation of anticipatory data-drift   condition to generate synthetic payload to test the model under such data distribution drift.

\item As the above requirements warrants generation of synthetic test data, it is important to generate the data based on some coverage criteria suitable for AI testing.

\item 
Use of synthetic test data to test fairness [18,44], robustness [7,28] properties in AI model which is majorly skipped by current industrial practices. 

\end{itemize}

Although existing techniques like GAN~\cite{GAN}, Variational Auto-encoder~\cite{VAE} can generate synthetic realistic data, they are  not customizable by user-specification. Moreover, existing  coverage criteria like neuron coverage, sign coverage, boundary value coverage, etc. ~\cite{DEEPXPLORE,DEEPGAUGE}  are not model-agnostic and cannot be combined with existing generative models.  

To address the majority of the above-mentioned drawbacks, we build a testing framework, called \system{}. The salient features \add{of} our framework are listed next.

\begin{itemize}[leftmargin=*]
\item First, to address the limited test data, we develop a technique to synthetically generate \emph{realistic} test data. For example, if there are two columns like \texttt{age} and \texttt{marital-status} - a random generation technique can generate married people younger than 20 even though such as behavior is not present in the training data.   Our data synthesis technique ensures that such a case never occurs. Specifically, the generated data (without any customization) has the similar statistical characteristics of the training data. 

\item Second, our synthetic data generation process is customizable. Consider a case where the training data is taken from a North America region where Male-Female ratio is, say 3:2, which is captured as a constraint inferred from the training data.  To cater to the testing for a different geographic region that has a different Male:Female (say 1:1), it is important to test the fairness metric with such synthetic data. \system{} allows to incorporate \emph{user-defined constraints} (UDC) to add and/or update the data constraints.  The  UDC is also important to capture any domain-based constraints which are hard to infer from the data or expected payload characteristics. The flexibility of having UDC is important from an AI Testing perspective to test various unprecedented what-if scenarios even before deploying the model in production. 

\cancel{
This also addresses an important problem of analyzing the model with anticipatory drift by generating synthetic test cases. It allows testing the model's behaviour under user-defined custom  specification already seen in some region of the input data which may possibly become a future trend and make the model fail badly. We envision that such a testing can help data scientist to perform testing before deploying the model in production and even perform 'what-if' testing on deployed models to cater for unseen possible future scenarios.
}
\cancelpc{
\item Third, to cater to the challenge of defining a model-agnostic coverage criterion, we re-use the notion of program path coverage. However, the notion of paths cannot be defined for all models.  Therefore, we introduce the notion of model-agnostic path-coverage and present algorithms for test-data synthesis ensuring high coverage. Essentially, we create a decision tree model with high fidelity which imitates the model under test. The test cases are then, equally distributed in the regions defined by the constraints in each path (hereafter called path-constrains) in the decision tree. Coverage of paths in the decision tree ensures decision region coverage of the model under test. 
}

\item \system{} performs goal-oriented test data synthesis. We perform group/individual fairness and robustness testing which are metamorphic properties~\cite{MT} whose testing does not require an oracle to get label for synthetic data. 

\end{itemize}
The current scope of our system is classification models for tabular data. To the best of our knowledge, no other techniques address this in tabular domain. Ribeiro et al. demonstrated the importance of testing NLP models in~\cite{CHECKLIST}. Our contributions are listed below:

\begin{itemize}[leftmargin=*]
\item  We develop an algorithm to generate realistic test data which is \emph{cutomizable} giving the user an opportunity to perform \emph{drift-testing}.

\item We define a \emph{model-agnostic  path coverage criteria} and re-use an existing global explainability algorithm to generate the data offering maximum coverage. 

\item  Our technique tests fairness and robustness with realistic synthetic data which is not done by other \emph{realistic} synthetic test data generation techniques. 
\cancel{
\item We implemented \system{} and experimented with the known benchmarks (used in IBM for testing) to demonstrate the effectiveness of our techniques.  

\item Our AI Testing system is already part of IGNITE Quality Platform~\cite{IGNITE} and is considered for inclusion in Watson OpenScale~\cite{WOS}.
}
\end{itemize}

\cancel{
Outline. The next section presents the relevant background knowledge and  Section~\ref{sec:algo} describes our test data synthesis algorithm in detail.
Section~\ref{sec:expt} presents the experimental results followed by related works in Section~\ref{sec:related}. Finally, we conclude with a brief summary along with future work in Section~\ref{sec:conc}.
}
\section{Background}
\label{sec:back}

In this section, we present the background of some of the already known concepts that we use in our framework.
\cancelpc{
\paragraph{Decision Tree Surrogate} 
We use an algorithm, called TREPAN~\cite{TREPAN} to create a decision tree surrogate of any given model with black-box access for defining its coverage.  Essentially, TREPAN uses training data with model predictions (instead of ground truth) to imitate the decision logic of the black-box model with a decision tree (an interpretable model).  TREPAN also generates random  synthetic samples when the number of training samples for an intermediate node in the decision tree is reduced below a given threshold. Note that it uses the labels from the target classifier even for those synthetically generated samples. The use of these additional samples helps to create a better surrogate. 
}
\paragraph{Fairness} 
A \emph{fair} classifier tries not to discriminate individuals or groups defined by the \emph{protected attribute} (like race, gender, caste, and religion)~\cite{testing-fse}. \eat{\cancel{A value of the protected attribute indicating a group that has historically been at a systematic advantage is the privileged group, while the rest is called an unprivileged one. In a binary classification scenario, the positive class corresponds to the favorable outcome that individuals wish to achieve, whereas the negative class corresponds to an unfavorable one.}} \eat{\cancel{Discrimination becomes objectionable/unfair when it places certain privileged groups at a systematic advantage and certain unprivileged groups at a systematic disadvantage.}} Next, we describe metrics related to group fairness (between two groups) and individual fairness (between two individuals) by using the following notations.  Each sample/individual is denoted as ($X$, $Y$, $Z$) where $X$ are all attributes used in the prediction, $Y$ is the corresponding ground-truths for the samples in $X$, and $Z$ is the binary protected attribute which may be included in $X$. A classifier is a mapping $h$: $X \rightarrow [0,1]$. The final prediction is denoted by $\hat{Y}$ where  $\hat Y = 1 \Leftrightarrow h(X) > \sigma$. We will use $P({\hat Y=1 }| Z = 1)$ as the probability of a favorable outcome ($\hat Y=1$) for the privileged group ($Z=1$). 

\noindent{\textbf{\textit{Group Fairness. }}} Below we recall some prominent group discrimination metrics that we use. Under the definition of disparate impact~\cite{DI}, a system is fair if:
 
\begin{scriptsize}	    $\dfrac{P(\hat Y=1 | Z = 0)}{P({\hat Y}=1 | Z = 1)} > \epsilon$. 
\end{scriptsize}
In other words, the probability of the favorable outcome of the unprivileged and privileged group should be more than a particular threshold.
Typically, based on US Govt. rules, in many scenarios $\epsilon = 0.8$. 
\eat{\cancel{Instead of ratio, some people prefer difference based operations. The metric demographic parity~\cite{DP} is defined below:

        }}

\noindent{\textbf{\textit{Individual Fairness. }}}
\eat{\cancel{
According to Dwork~\cite{FTA}, a system is individually fair if similar individuals receive similar treatments or outcomes. However, we take an alternative definition based on . }} Based on the notion of counterfactual fairness~\cite{CF},  a decision is fair towards an individual if it is the same in (a) the actual world and (b) a counterfactual world where the individual belonged to a different demographic group.  Essentially, a test case corresponding to individual fairness consists of a pair of samples where the two samples only differ in protected attribute values - one from the privileged group and the other from the unprivileged group.  Formally,
		$h(s) < \epsilon$ 
		and
		 $h(s') \geq \epsilon$ 
		 where $s.Z \neq s'.Z$
 and $s.y = s'.y~\forall y \in Y$. 
 
 Note that just removing the protected attribute from the training data doesn't ensure fairness in AI models due to the existence of possible indirect bias~\cite{testing-fse}, and therefore such a testing is required.
Further, our testing framework is generic enough to work for any other definition of fairness, but currently we are using the one used in ~\cite{THEMIS}~\cite{Aeqitas}~\cite{testing-fse}.

\eat{\cancel{
We implemented the algorithm~\cite{Aeqitas} for black-box testing of individual fairness. It has two phases - global search and local search. In the global search phase, it creates random samples and checks for the discrimination using the above formulation. For each discriminatory sample found in global phase, the algorithm executes a local search where it generates samples in the neighborhood of the discriminatory sample and checks the property. }}
\paragraph{Adversarial Robustness}  Testing for adversarial robustness consists of creating two realistic samples based on some perturbation function $p$  and checking if both of them give different outcomes (test failure). \eat{\cancel{Though adversarial robustness sounds very similar to individual fairness, it is actually different as there is no notion of fairness attribute here. The perturbation can alter with a valid value of any attribute. The resultant sample with perturbation should be close to the original one under some threshold and distance function (typically L2). }}

\section{Data Synthesis}
\label{sec:algo}

\begin{figure}[t]
\centering
\includegraphics[ width=7cm]{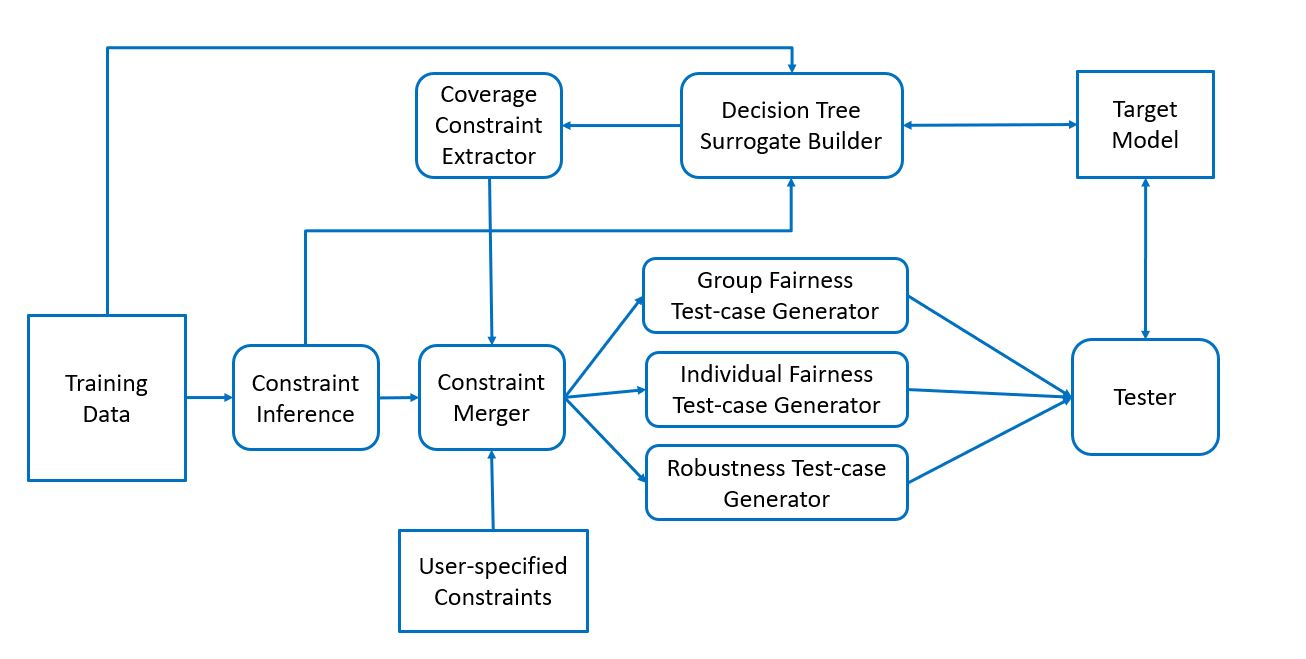}
\vspace{-4mm}
\caption{\system{} Architecture}
\label{fig:arch}
\vspace{-6mm}
\end{figure}

Figure~\ref{fig:arch} shows the different components of \system{} architecture and how they interact with each other. At first, \system{} processes the training data to obtain a set of constraints. Using such constraints, \system{} can generate synthetic data using a new constraint solver, after optionally merging  user-defined constraints and  path constraints. Finally, it generates the test cases for specific properties which are fed to testers. This section discusses different components of \system{} in detail.

\subsection{Constraints}
\label{sec:spec}
\newcommand{\cnn}{CAT-NUM-NUM }
\newcommand{\acatnum}{CAT-NUM }
\newcommand{\catnum}{\emph{cat\_num} }
\newcommand{\catcat}{\emph{cat\_cat} }
\newcommand{\catnumb}{\emph{cat\_num}}
\newcommand{\catcatb}{\emph{cat\_cat}}
\newcommand{\numnum}{NUM-NUM }
\newcommand{\ctext}{\textcolor{red}{TEXT} }
\newcommand{\num}{NUMERIC }
\newcommand{\float}{FLOAT }
\newcommand{\cattext}{CAT\_TEXT }
\newcommand{\acattext}{\textcolor{red}{CAT-TEXT} }
\newcommand{\cempty}{EMPTY }
\newcommand{\cdate}{DATE }
\newcommand{\datedate}{DATE-DATE }

The first step is to understand each column in the given data and find associations between different type of columns.  Note that we ignore the label column in this phase.
As per our constraint language specified in Table~\ref{tab:c}, constraints are of two types - \emph{column constraints} which are defined for each column, and \emph{association constraints} which are defined based on more than one column. 

\noindent{\textbf{\textit{Column Constraints}}}. We assume that a column can have either numeric or category datatype, the latter having a fixed set of unique values. Based on the column's datatype, different kinds of distribution constraints are inferred. For each category column, \system{} gathers frequency distribution of all the unique set of values, whereas for numeric column, \system{} gathers statistical properties such as minimum-maximum bound and various statistical distributions. 
Note that we try to fit five common statistical distributions for numeric columns - $uniform$, $normal$, $beta$, $exponential$ and $gamma$, using Scipy's distribution fit~\cite{scipy-2020} that uses maximum likelihood estimation technique, and then use the \textit{Kolmogrov Smirnov} (KS) test~\cite{kolmogrov,kstest} to check which distribution fits the column best. 
KS test compares the data with a reference probability distribution, and measures the distance  between empirical distribution function of the sample and cumulative distribution function of the reference distribution. The KS statistic value (or $p$-value) is high (1) when the fit is good, and low (0) otherwise. The lack of fit is significant if $p$-value $<$ $0.05$.
We say that a distribution fits the column if $p$-value from the KS test $>$ $0.05$ and is the best amongst the distributions checked. Due of this check, a numerical column may not have any associated distribution.

\begin{table}
\centering
\caption{Constraints}
\vspace{-4mm}
\begin{tabular}{|lll|}\hline
$constraints$ & ::=  & $column~~|~~association$\\
$column$ & ::= & $categorical~~|~~numerical$\\
$categorical$ & ::=  & $frequency\_distribution$\\
$numerical$ & ::= & $statistical~~|~~distributions$\\
$statistical$ & ::= & $min~~|~~~max$\\
$distributions$ & ::= & $normal~~|~~uniform~~|~~beta$\\ 
 & & $exponential~~|~~gamma$\\
$association$ & ::= & $cat\_cat~~|~~cat\_num$\\
$cat\_num$ & ::= & $\forall v \in categorical, (v, numerical)$\\
$cat\_cat$ & ::= & $\forall v \in categorical, (v, categorical)$\label{tab:c}
\\\hline
\end{tabular}
\vspace{-4mm}
\end{table}

\noindent {\textbf{\textit{Associations.}}} We aim to capture two types of relationships involving a pair of source and target column. 
The \catcat is defined between two category columns. We perform \textit{Chi-square}($\chi^2$) \textit{test}~\cite{chisquare} \add{and use uncertainty coefficient, Thiel’s U~\cite{theil1970estimation}} to measure independence between categorical attributes to identify the source and target column.
\add{Thiel’s U is based on the conditional entropy between two nominal attributes and measures the degree of association from the source to target, with the value in range $[0,1]$, where $0$ means no association and $1$ represents a full association.}
We capture the frequency distribution of values in the target column for each unique value-combination in the source column. 
The \catnum association is defined between a category source and a numeric target column.  For every unique value in the source category column, we find column-level constraints for the filtered samples in the target numeric column. For numeric target column, we do not represent joint categorical source columns in the association constraint as the count of numerical samples becomes very less for each value-combination of source categorical columns which prohibits distribution fit. 
However, our synthesis phase (described later) partially considers all the \catnum association constraints for a numerical column. 

Note that the datasets are generally preprocessed before training a model and therefore, all the redundant columns or columns dependent on other columns are pruned. Hence, an association from a numeric column to another numeric one capturing a polynomial relationship may not be useful, especially for testing ML models. Furthermore, it is also possible to define many other constraints, for example, associations defined by considering more than two columns. But, we found the suggested set of constraints enough and best suited for our data synthesis technique.

To capture the dependency between all the constraints, we define a directed graph, $G(N,E)$ and call it the \emph{Constraint Dependency Graph} ($CDG$). Each node $n\in N$ in this $CDG$ corresponds to a feature/column 
and is annotated with an associated inference error related to individual feature constraints. This $inf\_error(n)$ is $0$ for categorical columns and numerical columns where no distribution is inferred. For a numerical column $n$ having some distribution, $inf\_error(n)=1-p$-value. Each directed edge $e \in E$ in the graph corresponds to either a \catnum association or part of \catcat association between source and target node. Edge $e$ is annotated with $inf\_error(e)$ which is \add{\cancel{$0$}} \add{$1-u$-value} for \catcat edges, \add{where $u$-value is the uncertainty coefficient for $e$,}
and for each \catnum edge, it is an average of numerical distributions errors for all category source values. Since a numerical column may not have any  continuous distribution function in feature constraint (but may have association), each node is  classified into two types - 1) \texttt{GenNode}: one having distribution function and 2) \texttt{NonGenNode}: not having any distribution function. Later, we demonstrate how to use this $CDG$ for data synthesis.
%
\eat{\cancel{
Similar to every node, each $CDG$ edge $e$ has an inference error for related association i.e $inf\_error(e)$.
Note that \texttt{CAT-CAT} has zero inference error. . 
}}

\eat{\cancel{
This methodology for finding distributions can be extended to any standard distribution. We check for only five distributions since these are the most common ones and fit to most of the real data. Some of these, like beta distribution, actually capture a family of distributions.}}
\eat{\cancel{
min, max, mean and distribution of values in the target column. We also find error in this constraint which depicts how good fit this distribution is on the target column data, and return it as inference error.}}
%
\eat{\cancel{
The \datedate association is a straightforward one where we find the minimum and maximum difference between two date columns. 
This gives information on the range of target column w.r.t the source column. 
For example, if both the minimum and maximum difference for two date columns are positive, we know that the target date is always later than the source date, and we also know the range of values target date can take w.r.t. the source date.
A simple example for this association is the difference between order date and delivery date for a product. In this case, the association conveys that the delivery date is at least min-diff days after order date, but at most max-diff days after the order date.}}
\eat{\cancel{
For the \numnum association, we try to find a polynomial function from the source column to the target column. The target column may not be an exact function of the source column, but an approximate one. So we also find the error of how good fit this polynomial is, and return it as inference error. }}
%
\eat{\cancel{
We try to fit a polynomial of degree $d$ (initially $d=1$) from the source column to the target column, and if the fit is not good enough then try to fit a polynomial of degree $d+1$. We say that the fit is good enough if the normalized root mean square error between the target column and the target column computed using the polynomial is less than a threshold ($0.01$).
We try to fit polynomials only till degree $4$ as it is a computationally expensive operation, and also more practical for most real datasets.
We use the numpy's polyfit functionality~\cite{numpy} to fit a polynomial from the source column to the target column. This method uses least square fit method to fit the polynomial and returns the parameters for the corresponding degree.}
 
\cancel{In this case, the polynomial associations may not be inferred, which could have been inferred from the original data.}}

\eat{\cancel{Most of the associations (even for same type of source and target) described above are assymetrical, i.e., existence of an association from a column $c_1$ to a column $c_2$ does not necessarily mean existence of similar association from $c_2$ to $c_1$.  In fact, the reverse association does not exist in many cases.}
\cancel{For example, if there exists a polynomial of degree $d$ ($d>1$) from a column $c_1$ to a column $c_2$, there exists no polynomial from $c_2$ to $c_1$. For a pair ($c_1,c_2$) of columns, we find association from $c_1$ to $c_2$ and if we do not find any association in this direction, we find association from $c_2$ to $c_1$. }}
%
%




Consider a toy dataset with five categorical attributes (\texttt{gender, education, martial, age-grp, intelligence}) and one numeric attribute (\texttt{salary}) with relevant associations between them. An exemplar $CDG$ is shown in Figure~\ref{fig:graph} (G1).

\begin{figure}[t]
\centering
\includegraphics[width=8cm]{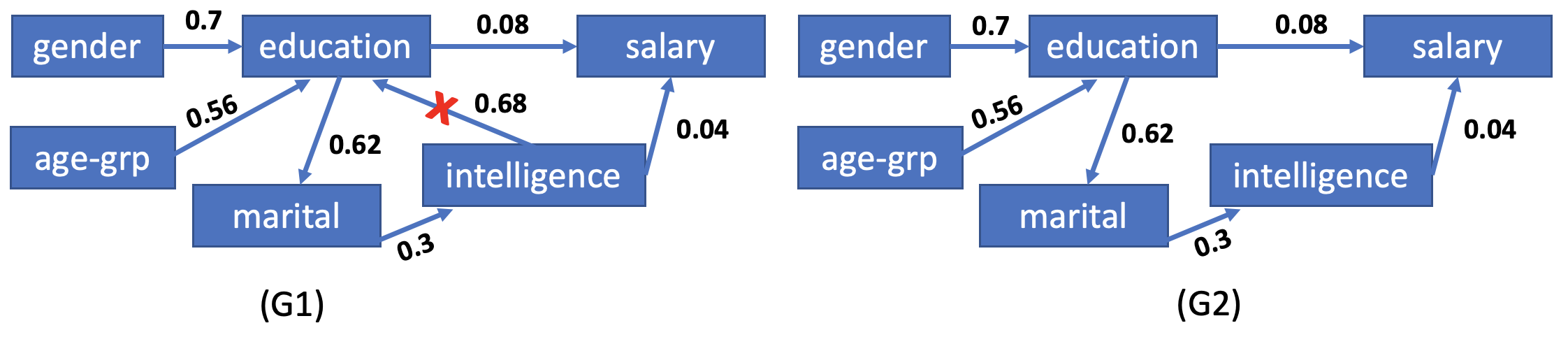}
\vspace{-4mm}
\caption{CDG (G1) and DAG (G2)}
\label{fig:graph}
\vspace{-4mm}
\end{figure}


\subsection{User-Defined Constraint Specification}
\label{sec:user-defined}
Apart from the data constraints, \system{} also enables a user to add new constraints or delete/modify the existing data constraints in form of \emph{user-defined} ones (UDC) which take precedence over the existing data constraints.
Our specification supports three types of UDCs as listed below.

\begin{itemize}[leftmargin=*]
    \item \textit{Add UDC} -  User can add 
    constraints which spans both association and column related constraints. For example, if data constraints infer that values of salary lie in the range [2k-30k] with no distribution, then, using UDC, user can override this random generation by specifying, say a normal distribution with loc=15k and scale=0.1 lying in the same bounds [2k-30k].
    \item \textit{Modify UDC} - User can specify constraints to modify/override the existing data constraints, both association and feature related constraints, as in the usage scenarios below.\\
    \emph{Range Modification} - Let's consider that the user wants to generate salary values for females in a particular range, say, 5k-50k, which is different from the bounds 2k-30k as specified in \catnum
    association between gender and salary. Using user-defined constraints, he/she can override the value bounds or statistical distribution in feature/associations constraints with the desired values. \\
    \emph{Distribution Modification} - Another scenario of its utility can be when the user wants to generate test cases to check for group fairness in AI model. For a protected attribute, say gender, the frequency distribution of M:F is, say 3:1, in the feature constraint. But, it may be desired to check for fairness with the test set having a different relative proportion of M and F, say 1:1. Our UDC template allows such frequency distribution overriding.
    \item \textit{Delete UDC} - User can ask the system to drop certain associations or feature constraints from the inferred data constraints during test data synthesis. For instance, drop range constraints on an attribute, or drop associations between a protected (gender) and non-protected (salary) attribute during test data generation.
\end{itemize}
Later in Section~\ref{sec:Synth}, we discuss the way to handle such UDCs in our synthesis algorithm.

\subsection{Realistic Data Synthesis}
\label{sec:Synth}

\begin{algorithm}
\scriptsize
\SetAlgoLined
\SetAlgoLined\DontPrintSemicolon

\SetKwProg{Fn}{Function}{}{end}
\Fn{DataSynthesis(InputData, n)}{
    DataConstraints = genConstrains(InputData)\\
    CDG =  createCDG(DataConstraints)\\
    return DataSynWithDataCons(CDG, n)\\
}

\Fn{createCDG(DC, n)}{
    G.nodes = \{\}\\
    G.edges = \{\}\\
    \ForEach{$c \in DC.columns$}{
        G.nodes += $n_c$
    }    
    \ForEach{$fc \in DC.feature\_constraints$}{
        $n_c.constraints$ += $fc$\\ 
        \uIf{($n_c.type == NUM \wedge \exists c \in fc~s.t.~c.type==cont\_distribution$)}{
            $n_c.GenNode = False$
        }\Else{
            $n_c.GenNode = True$
        }
    }    
    \ForEach{$ac \in DC.assoc\_constraints$}{
        $n_c.asso\_constraints$ += $ac$ 
    }
    return G
}
\Fn{DataSynWithDataCons(CDG, n)}{
    DAG = preProcess(CDG)\\
	return generate(DAG, n)			
}
\Fn{preProcess(CDG)}{
    DAG = CDG.copy()\\
    \While{$\exists~directed~cycle~$c$ \in DAG$}{
        delete edge ($n_i,n_j$) with minimal confidence s.t. $n_j.GenNode = True$
    }
    return DAG
}

\Fn{DataSynWithDataCons(DAG, n)}{
    \While{$n_j \in topologicalOrder(DAG.nodes)$}{
        data = genDataEachColumn($n_j$, DAG, n)\\
        $n_j.data = data$
    }
}
\Fn{genDataEachColumn($n_j$, DAG, n)}{
        \uIf{($n_j.GenNode \wedge n_j.type == NUM$)}{
            dataOwn = gen($n_j$.dist, n, $n_j$.range, $n_j$.isUnique)\\
            minDiv = +Inf\\
            minDivData = null\\
            \ForEach{(e = ($n_i$,$n_j$)}{
                data=genData(e,$n_i$)\\          
                div = KL-div(dataOwn, data) \\    
                \If{$div < minDiv$}{
                    minDivData = data\\
                }
            }
            $n_j.data = data$\\
        }\uElseIf{($n_j.GenNode$)}{ \tcp*[l]{type CAT}
            e = ($n_i$,$n_j$) for any $n_i$\\
            data=genData(e,$n_i$)\\          
            $n_j.data = data$\\
        }\Else{\tcp*[l]{NonGenNode}
            e = ($n_i$,$n_j$) s.t. $error(e)$ is min for all $n_i$\\
            data=genData(e,$n_i$)\\          
            $n_j.data = data$\\
        }
    
}

\caption{Data Synthesis using Data Constraints}
\label{algo:ds}
\end{algorithm}

In this section, we present the overall algorithm of data synthesis in three stages - starting from realistic data synthesis from data constraints, and subsequently adding user-defined constraints and coverage constraints. Furthermore, the synthesizer also expects the count of samples to be generated as an input. Note that the synthesizer does not generate data for the label column.


In order to generate values for the toy example, we need to consider the following set of constraints.
\begin{enumerate}[leftmargin=*]
    \item Feature Constraints of individual attributes
    \begin{itemize} [leftmargin=*]
        \item \emph{Continuous Marginal Distribution}:        Synthesizer can generate values by sampling from the existing distribution e.g. \texttt{salary} from normal distribution.
        \item \emph{Discrete Marginal Distribution}: Synthesizer can generate required number of values for a fixed frequency ratio (e.g. \texttt{gender} in a, say M:F=2:1) using enumeration.

\eat{\cancel{
    \item \emph{Range and Difference Constraints} (\texttt{DATE}):\\ Synthesizer can generate required number of values using enumeration within the range.
        }}
    \end{itemize}
    \item Association Constraints
            \begin{itemize}[leftmargin=*]

    \eat{\cancel{
            \item \emph{Polynomial Relationship} (\texttt{NUM-NUM} and \texttt{DATE-DATE}):
            To generate values for \texttt{age} and \texttt{salary} such that they satisfy a polynomial equation along with their upper and lower bounds, synthesizer can use any existing constraint solver.
            }}
            \item \emph{Conditional Distribution for Numerical variables} (\catnumb):
            Using sampling from the  distribution for specific categorical values. For \texttt{education=masters}, if \texttt{salary} follows a uniform distribution with specific bounds, then for every \texttt{masters} value, we can sample a value from this distribution for \texttt{salary}.
            \item \emph{Conditional Distribution for Categorical variables} (\catcatb):
            For each value-combination of \texttt{gender},   \texttt{age-grp} in source,  say  \texttt{\{Female, Senior\}}, the frequency distribution of target \texttt{education} (say primary:secondary:tertiary=1:2:3) is used to generate synthetic values for \texttt{education} for all rows containing \texttt{\{Female, Senior\}}. 
        \end{itemize}
\end{enumerate}

But, the challenge is how to generate values for an attribute while considering multiple constraints at once. \cancel{, especially in the presence of errors associated with distribution fittings and association relationships. }

Data synthesis for a node can be done in two ways -  either by using the generation procedure for feature distributions (marginal distributions) or by solving the association constraint in each incoming edge whose source values have already been generated (conditional distribution).  This essentially outlines the challenges in our data synthesis algorithm - 1) the order of processing the nodes through $CDG$, and 2) how to compute the data for each node.   

Note that the $CDG$ can contain cycles involving three or more nodes. To address the node processing problem, the synthesizer pre-processes the $CDG$ to remove any cycles, essentially turning it into a directed acyclic graph (DAG) (Graph \texttt{G2} in Figure~\ref{fig:graph}). The synthesizer breaks a cycle by removing \emph{any one incoming edge to a \texttt{GenNode} with maximal error in the cycle}. This is because a \texttt{NonGenNode} requires an association constraint to generate its value and removing the most-erroneous edge ensures that less error is propagated through the graph. Once the DAG is obtained, the generation proceeds node-by-node in topological order or bottom-up order in the DAG. 
   
Assuming that there is no singleton  disconnected node in the graph, this transformation leads to a DAG which has only \texttt{GenNode}s in the leaves. The synthesizer generates values in the leaves using the marginal distribution-based generation procedure  described before. Subsequently, the generation proceeds through the DAG in the topological order starting from the leaves, generating values for each non-leaf node.

Consider a case, where a non-leaf node has a single incoming edge. In this case, it is easy to see that considering the association constraints to compute the data (as described before) is better with respect to its own distribution as it incurs zero loss for categorical columns and minimal distribution loss for the numeric  columns while also respecting the relationship with the other feature values.  


In the case of multiple incoming edges, there are three cases. If the current node $n$ is a \texttt{GenNode} and numeric, then for every incoming edge $e$ (of type \catnumb), the synthesizer generates data and computes KL-divergence error~\cite{kldivergence} ($kl\_error(e)$) with respect to its own distribution in column constraints. It finally selects the edge $e'$ having minimal kl\_error. Based on ($e'=(cat,n)$), it samples data from the numerical distribution corresponding to every categorical value of variable $cat$. For each such generated value of $n$, it accepts the samples only if  ($\forall e''=(cat',n), e''!=e$) the range constraints for $e''$ are also satisfied, thereby minimizing the difference from his own distribution while maintaining cross-feature relationships. For example, \texttt{salary} is generated by the distribution corresponding to each \texttt{education} value, but the generated value also respects the \texttt{salary} range of the corresponding \texttt{intelligence} value. If this node $n$  is a \texttt{GenNode} but categorical, it uses \catcat relationship involving multiple source nodes to compute the data. In the third case, i.e. for a \texttt{NonGenNode}, the synthesizer selects the incoming edge with minimum error ($error(e)$) and computes the data. We define error of node $n$ and edge $e$ as follows. 
\[
error(n) = \begin{cases}
inf\_error(n), \text{if $n$ is a leaf-node}, \\ 
 error(e') + kl\_error(e') \times inf\_error(n), \text{ow.}
 \end{cases}
\]
where $error(e) = inf\_error(e) + error(n')$ when $n'$ is source of $e$, and $e'$ is the incoming edge selected to generate values of node $n$.
Essentially, this synthesis error of node $n$ is  propagated to all the outgoing edges in the graph.
The pseudo-code of our realistic data synthesis algorithm is presented in Algorithm~\ref{algo:ds}.


In the above example, the order of processing nodes is \texttt{gender},
\texttt{age-grp}, \texttt{education}, \texttt{marital}, 
\texttt{intelligence}, 
and \texttt{salary}. 
The synthesizer applies the case of multiple incoming edges for the \texttt{education} and \texttt{salary} nodes.

\begin{algorithm}
\scriptsize
\SetAlgoLined
\SetAlgoLined\DontPrintSemicolon

\SetKwProg{Fn}{Function}{}{end}

\Fn{DataSynWitUDC(CDG, n, UDC)}{
    DAG = preProcessWithUDC(CDG, UDC)\\
	return DataSynWithDataCons(DAG, n)			
}
\Fn{preProcessWithUDC(CDG, UDC)}{
    DAG = CDG.copy()\\
    DAG = processUDC(DAG, UDC)\\
    \While{$\exists~directed~cycle~c \in DAG$}{
        delete edge $e=(n_i,n_j)$ with minimal confidence s.t. $n_j.GenNode = True$
        where 
    }
    return DAG
}
\Fn{processUDC(DAG, UDC)}{
    \ForEach{$c \in UDC.Add$}{
        delete $c$ from $DAG$\\
    }
    \ForEach{$c \in UDC.Add$}{
        \If{$c=n_c.cons$ is a feature constraint}{
            Add to the corresponding node $n_c$ in $DAG$\\
            remove incoming edges to $n_c$ 
        }
        \If{$c=(n_i,n_j).cons$ is an association constraint}{
            Add to $(n_i,n_j).cons$ to $DAG$\\
            remove other incoming edges to $n_j$ \\
        }
    }
    \ForEach{$c \in UDC.Modify$}{
        
        \uIf{$c=n_c.cons$ is a feature constraint $\wedge$   of type distribution}{
            Modify the corresponding node $n_c$ in $DAG$\\
            remove incoming edges to $n_c$ 
        }\uElseIf{$c=(n_i,n_j).cons$ is an association constraint $\wedge$}{
            Modify $(n_i,n_j).cons$ im $DAG$\\
            remove other incoming edges to $n_j$ \\
        }\Else{
            $c=n_c.cons$//
            Update constraints in node $n_c$ DAG// 
            Make it high confidence
        }
    }
    return DAG
}

\caption{Data Synthesis using Data Constraints and User-defined constraints}
\label{algo:ds-udc}
\end{algorithm}

\paragraph{Handling User-Defined Constraints (UDCs)}
If user specifies some additional constraints to override the existing ones obtained from data, then the synthesizer first merges the two set of constraints.
This resultant merged constraint set is then used to synthesize test data by a modification of the above process, as shown in Algorithm~\ref{algo:ds-udc}. Essentially, UDC nodes are made leaf node in the DAG by removing all the incoming edges. Also, the effect of UDC generated data is propagated through the DAG in such a way that it overrides any incoming edge selection choice described above.    

UDCs are merged with data constraints with the existing data constraints, as below. Note that the UDCs have higher priority than the inferred data constraints.
\begin{itemize}[leftmargin=*]
    \item \textit{Add/Modify UDCs} - For every such UDC, we add/modify that constraint in our set of data constraints. In case of conflict, we record the over-riding constraints in our data constraints and make sure that the former get precedence over the latter during the synthesis process. \\
    - \emph{Range modification} - Note that there is a need to handle the cases of range modification in a more systematic way. Consider a case where data constraints have a feature constraints on an attribute, say salary, where the values are bounded by range $(min_{data}, max_{data})$, say 2k-30k and follow a uniform distribution with certain parameters. Now, user specifies to override just the range for Salary as $(min_{user}, max_{user})$, say 5k-50k without any change in associated statistical distribution. Overriding just the bounds in the data constraints may impact the present distribution parameters. Therefore, using the original min-max bounds 2k-30k along with the specified uniform distribution parameters in the data constraints, we first generate a set of salary values, V. We then scale these generated values using the below formula for every value v $\in$ V to the final output values for salary. 
    \[
    v_{scaled} = min_{user} + \frac{(v - min_{data})*(max_{user}-min_{user})}{(max_{data}-min_{data})}
    \]
    This scaling ensures that the mean of scaled values is linearly scaled version of the values before scaling.
    \item \textit{Delete UDC} - If the overriding constraints specify to drop an entire feature, $f$ from the resultant test set, then we drop the column constraints along with all its related association constraints where $f$ is mentioned as either source or target.
\end{itemize}

\cancelpc{
\paragraph{Path Coverage Constraints}

The key idea here is to build a decision tree of the target model (in a model agnostic way) and use its path coverage as the coverage criteria for the target model. \system{} uses the TREPAN algorithm~\cite{TREPAN} to create a surrogate decision tree (see Section~\ref{sec:back}). TREPAN generates random data to augment the input training samples. We essentially change this step to generate realistic data instead of using random data. All path constraints are fetched for each path. The aim is to generate data satisfying the path constraints and satisfying majority of the data and UDC.  Below, we describe the change in the above procedure to generate  $n$ samples belonging to a path.

Each constraint in a path is either an equality predicate for categorical value or range constraints for numeric values. The columns with equality predicate are made leaf node in the CDG by deleting all incoming edges. However, the range constraints are added to the CDG nodes as additional feature constraints. Similar DAG generation and topological generation order starts with one necessary modification - only in the case where a range constraint does not agree with the association constraints, then the association constraints are ignored and the data is generated by considering the feature and range constraints. 
}

\remove{
\subsection{Realistic Data Synthesis}
\label{sec:Synth}
This subsection discusses our overall approach to synthesize \textit{realistic} data using constraints inferred by \emph{Constraint Inference} engine (see Section~\ref{sec:spec}). \system{}'s synthesizer expects two inputs: \textit{constraints} - data constraints for an input data, and \textit{n} - number of required synthetic tests to be generated.

Consider a toy dataset with four attributes (\texttt{gender(CAT\_TEXT), salary(FLOAT), age(NUMERIC), education(CAT\_TEXT)} with the following types of associations between them.
\begin{center}
\scriptsize
\begin{tabular}{cccc}
assoc-id & src & tgt & assoc-type \\
\hline
a1 & gender & age & CAT-NUM \\ 
a2 & age & salary & NUM-NUM \\
a3 & gender & salary & CAT-NUM \\
a4 & gender & education & CAT-CAT\\
a5 & education & gender & CAT-CAT\\
\hline
\caption{Associations}
\end{tabular}
\end{center}\\ 

In order to generate values for this example, we need to consider the following set of constraints.
\begin{enumerate}[leftmargin=*]
    \item Feature Constraints of individual attributes
    \begin{itemize} [leftmargin=*]
        \item \emph{Continuous Distribution} in case of \texttt{NUMERIC} and \texttt{FLOAT}\\
        If values of \texttt{age} follow a normal distribution with specific bounds, we can sample required number of age values from this distribution using existing techniques.
        \item \emph{Discrete Distribution} followed by \texttt{CAT\_TEXT} and \texttt{CAT\_NUM}\\ Generation of values for \texttt{gender} in a fixed frequency ratio, say M:F=2:1, can done by using an existing solver.
    \end{itemize}
    \item Association Constraints
        \begin{itemize}
            \item \emph{Polynomial Relationship} in case of \texttt{NUM-NUM} and \texttt{DATE-DATE}\\
            To generate values for \texttt{age} and \texttt{salary} such that they satisfy a polynomial equation along with their upper and lower bounds, we can use any existing constraint solver.
            \item \emph{Joint Continuous Distribution} captured by \texttt{CAT-NUM} \\
            For \texttt{gender=M}, if salary follows a uniform distribution with specific bounds, then for every \texttt{M} value, we can sample a value from this distribution for \texttt{salary}.
            \item \emph{Joint Discrete Distribution} in case of \texttt{CAT-CAT} \\
            For \texttt{gender=F}, if frequency distribution of \texttt{education} values, say masters:primary=20\%:80\%, then we first fetch the total count of \texttt{F} values in \texttt{gender}, say $10$. Therefore, $2$ females are assigned masters while $8$ are assigned primary as \texttt{education}.
        \end{itemize}
\end{enumerate}

But, the interesting part lies in how we try to generate values for an attribute while considering multiple constraints at once.

\paragraph{Dependent Features.} If a feature is specified as a target in any association in the fetched set of constraints, only then it's classified as dependent. As per our toy example, only \texttt{gender} is independent.

\paragraph{Feature Process Order.} We employ a column-wise generation scheme, where we generate data for independent features prior to the dependent ones. This ensures that at the time of generating column values for a dependent feature, the required source feature values have already been generated and are ready to be consumed. Therefore, the source and target features in all association constraints are analyzed, thereby resulting an order in which values for different features are to be generated. \\
\emph{Associations Pruning.} The process starts with creating a graph where all features are represented as nodes, and for every association constraint between source \textit{src} and target feature \textit{tgt}, a directed edge from node \textit{src} to \textit{tgt} is added in this graph. The graph then undergoes pre-processing to prune edges/associations in this graph.
G1 in Figure~\ref{fig:graph} shows the association graph created for our exemplar dataset. Here, associations a4 and a5 form a cycle in the graph. But, in order to find the feature process order, we need an acyclic graph. Therefore, we \emph{break the existing cycles} or self loops in the graph by removing the last added edge/association in that cycle (i.e. a5). The resultant acyclic graph G2 is further looked for the cases where multiple transitive paths exist between two nodes, such as path (a1, a2) and (a3) between \texttt{gender} and \texttt{salary}. We \emph{remove such transitivity} by considering only the path with the shortest length and breaking the rest. Association a2 is therefore removed from graph G2 to yield final pruned set of association graph G3. The intuition behind considering shorter paths is to make sure that minimal approximation seeps from the source values to the target values while processing associations.
We then \emph{topologically sort} this directed acyclic graph G3 to return the order to process and generate values for the features, which in our case is \texttt{gender} followed by (\texttt{age, salary and education} in any order).
\begin{figure}[t]
\centering
\includegraphics[width=8cm]{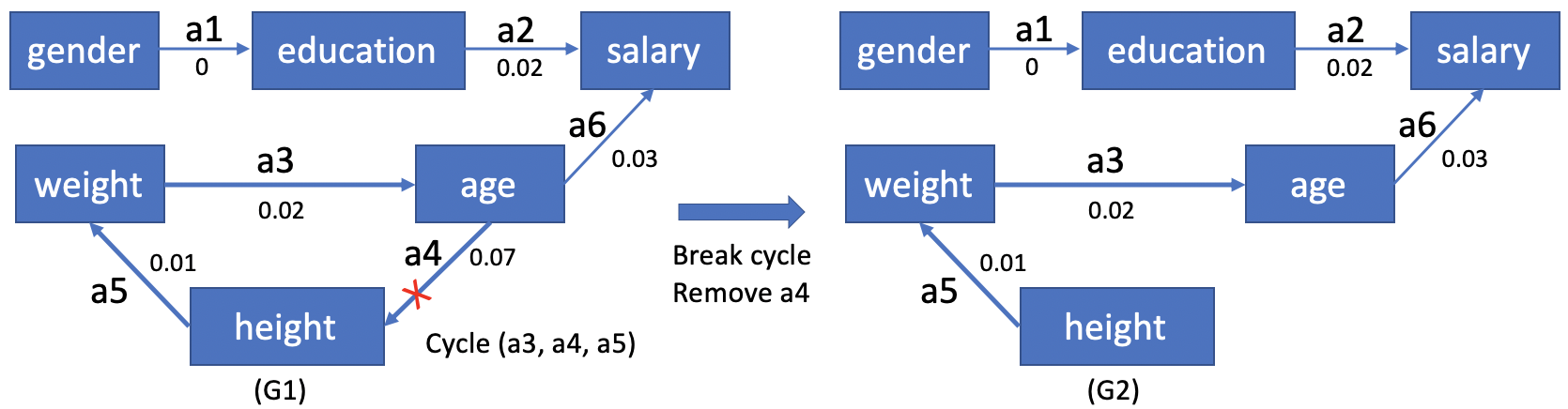}
\caption{Associations Pruning}
\label{fig:graph}
\end{figure}
\paragraph{Generate by Feature Constraints.} Next, we discuss how to generate values for an attribute by satisfying just the individual feature constraints. No associations are considered here during the processing. As per the process order, \texttt{gender} is independent and hence its values will be generated by using feature constraints only. These values of \texttt{gender} will later act as source values while generating values for all its direct dependents using associations. Based on the datatype of the input feature, we generate values as follows.
\begin{itemize}[leftmargin=*]
    \item \textit{NUMERIC/FLOAT}: We generate the values for NUMERIC and FLOAT datatypes by drawing samples from the given statistical distribution in the feature constraint. Note that apart from distribution name and parameters, we also consider minimum and maximum bounds along with the uniqueness constraints. In case where no distribution was inferred for the feature in the constraints, we use \emph{Uniform} as the default.
    \item \textit{CAT\_TEXT/CAT\_NUM}: For category-based datatype, the constraints report relative frequency distribution of different possible category values. We generate the values for a categorical feature such that they follow a similar relative frequency distribution as specified in the constraints. In case of unavailability of such constraint, we generate values by randomly picking categories from the possible value set.
    \item \textit{DATE}: We generate random date values between permissible start and end values for features with DATE datatype. Note that while generating values, we do consider uniqueness constraints and output the values in the same date format as the one specified in the feature constraints.
\end{itemize}

\paragraph{Generate by Associations.} For every dependent feature $f$, we begin with fetching all the association constraints with $f$ as target. Here, the discussion concerns with processing and satisfying this subset of associations along with the feature constraints depending on the datatype of target feature $f$, as described next.
\begin{itemize}[leftmargin=*]
    \item \textit{NUMERIC/FLOAT}: Values of such data types can either be generated by constraint-solving or generative approach in case of \texttt{NUM-NUM} and \texttt{CAT-NUM} associations respectively. If \texttt{NUM-NUM} exists, then we first try to generate target values using \texttt{NUM-NUM} associations which are then validated against applicable distributions in \texttt{CAT-NUM} and \texttt{NUM} feature constraints. In case of existence of only \texttt{CAT-NUM} association, we sample a target value using one of the applicable distribution and validate if it fits the remaining \texttt{CAT-NUM}s and feature constraint. In case the validation of any generated value fails, we generate value using a fallback mechanism as per the following priority - \texttt{NUM-NUM $>$ CAT-NUM $>$ NUM} feature constraints. We have used z3-solver~\cite{Z3} to solve polynomial equations for \texttt{NUM-NUM} associations.\\
    \emph{Numeric Validator} - We use the same Kolmogrov Smirnov(KS) test [28,35] used in Section~\ref{sec:spec} to check the goodness of fit of generated target values against desired statistical distribution. If the resultant $p$-value from the KS test is greater than a set threshold (in our case $0.05$), only then we accept the fit.
    
    \item \textit{CAT\_TEXT/CAT\_NUM}: If an association of type \texttt{CAT-CAT} exists between two attributes, say education (source) and gender (target) such that for masters, male:female = 3:1, while for primary education, male:female = 1:1. Considering masters:primary = 1:1 in already generated 8 source values, we generate 3 (male, masters), 1 (female, masters), 2 (male, primary), 2 (female, primary) pairs in the synthetic tests. 
    \item \textit{DATE}: \texttt{DATE\_DATE} is the only type of association related to this datatype which is satisfied along with DATE related feature constraints simultaneously to generate data. Let's consider an exemplar dataset with two attributes OrderDate and DeliveryDate binded by the following constraints. \\
    \texttt{DeliveryDate = OrderDate + [2-10days]} (DATE-DATE)\\
    \texttt{15-05-1995 $\leq$ DeliveryDate $\leq$ Today} (Feature constraint) \\
    All these constraints for DeliveryDate are fed to a solver along with already generated values of OrderDate to yield DeliveryDate values which satisfy all these conditions.
\end{itemize}
\paragraph{Generate by considering User-Defined Constraints (UDCs)}
If user specifies some additional constraints to override the existing ones in data, then we first need to merge the two set of constraints, so that the resultant set comprises of non-conflicting constraints. This resultant merged constraint set is then used to synthesize test data by following the same process as described before.
UDCs are merged with data constraints by handling their conflicts  with the existing data constraints. Note that the UDCs have higher priority in case of any conflict with inferred data constraints. Following is our strategy to handle different types of UDCs. 

   \begin{itemize}[leftmargin=*]
    \item \textit{Add/Modify UDCs} - For every such UDC, we add/modify that constraint in our set of data constraints. In case of conflict, we record the over-riding constraints in our data constraints and make sure that the former get precedence over the latter during the synthesis process. \\
    - \emph{Range modification} -  Note that there is a need to handle the cases of range modification in a more systematic way. Consider a case where data constraints have a feature constraints on an attribute, say salary, where the values are bounded by range $(min_{data}, max_{data})$, say 2k-30k and follow a uniform distribution with certain parameters. Now, user specifies to override just the range for Salary as $(min_{user}, max_{user})$, say 5k-50k without any change in associated statistical distribution. Overriding just the bounds in the data constraints may impact the present distribution parameters. Therefore, using the original min-max bounds 2k-30k along with the specified uniform distribution parameters in the data constraints, we first generate a set of salary values, V. We then scale these generated values using the below formula for every value v $\in$ V to the final output values for salary. \\
    $
    v_{scaled} = min_{user} + \frac{(v - min_{data})*(max_{user}-min_{user})}{(max_{data}-min_{data})}
    $ \\
    This scaling ensures that the mean of scaled values is linearly scaled version of the values before scaling.
    \item \textit{Delete UDC} - We drop/disable such constraint from our set of data constraints. If the overriding constraints specify to drop an entire feature, $f$ from the resultant test set, then we drop the feature constraints along with all its related association constraints where $f$ is mentioned as either source or target.
\end{itemize}
}

\begin{algorithm}
\scriptsize
\SetAlgoLined
\SetAlgoLined\DontPrintSemicolon

\SetKwProg{Fn}{Function}{}{end}

\Fn{accuracy(n, CDG, UDC, DT, Model)}{
	   data = DS(CDG, n, DT, UDC)\\
	   label = labeler(data)\\
	   label1 = test(data)\\
	   return accuracy(label,label1)\\
}
\Fn{group\_fairness(model, n, CDG, UDC, DT, protected\_attribute, priv\_expr, fav\_outcome, metrics)}{
        result = []\\
	    data = DS(CDG, n, DT, UDC)\\
	    label1 = test(data, model)\\		   
	    \ForEach{$metric \in metrics$}{
	    	metric,value = evaluate(metric, data, label, protected\_attribute, priv\_expr, fav\_outcome)\\					add (metric,value) to result\\
	    }
	    return result\\
}
	
\Fn{individual\_fairness(model, n, CDG, UDC, DT, protected\_attribute, priv\_expr, fav\_outcome)}{
	    data = DS(CDG, n, DT, UDC)\\
	    indiv\_algo(data, model,protected\_attribute, priv\_expr, fav\_outcome, DS) \\ \tcp*[l]{indiv\_algo can be any off-the-shelf algorithm to detect individual discrimination}
}
	
\Fn{robustness(model, n, CDG, UDC, DT)}{
	    failure\_count = total\_failure\_count = 0\\	
	    data = DS(CDG, n, DT, UDC)\\
	    labels = test(data, model)\\
	    \ForEach{sample $\in$ data}{
	    	data1 = perturb(sample)\\
		    labels1 = test(data1, model)\\
            flag=0\\
		    \ForEach{label $\in$ labels1}{
		     \If{labels[sample]!=label}{
 		        \If{!flag}{	
		      	    failure\_count++;\\
             	    flag = 1\\      	
		        }
		        total\_failure\_count++;
		      }
		    }
		 }
	    return total\_failure\_count, failure\_count/|data|\\
}    
\caption{Property-based Test Data Synthesis}
\label{algo:ds-prop}
\end{algorithm}

\subsection{Property-based Test Data Synthesis}
\label{subsec:prop-test-synth}
We further intend to generate realistic test cases to test an input AI model for a number of properties, namely \emph{Individual Fairness}, \emph{Group Fairness} and \emph{Robustness}. \eat{Note that, all these properties are metamorphic in nature which do not require any oracle to tell the gold standard label for the synthetic test data.} We use the same base approach as presented above with a few changes depending on the property, as shown in Algorithm~\ref{algo:ds-prop}.
\eat{
For property-based testing, \system{} takes data constraints, required test size, class label and protected attribute as an input.}

The property-based test input generation starts with removing the feature related to input \textit{class label} along with all the association constraints where this feature is specified as either a source or target. This ensures that any approximation caused due to constraint inference or synthesis does not seep into the prediction labels for these test inputs.
\eat{\cancel{
\subsubsection{Correctness Test Case Generator}
\label{subsec:accuracy}
Since we generate test inputs conforming to the behaviour/constraints inherent in the input training data, therefore, these realistic test inputs, of reasonably good size, should be able to cover most of the paths of the trained model. Hence, they can be put to test correctness of an input AI model too. As mentioned earlier, we don't use constraints to synthesize prediction values for these realistic test cases. Label assignment ideally should be a done by manual intervention, but, we instead use a \emph{Multi-model Label Generator} to guess labels for these inputs. \\

\paragraph{\textit{Multi-model Label Generator.}} We train multiple classifiers on the input data and choose the best label for test data based on the prediction and confidence of all the classifiers.
For each test input, we aggregate the confidence value on all the possible labels by all the classifiers, and then choose the label with highest aggregate.
We have experimented with other strategies like taking majority of predictions from all classifiers, taking majority of predictions  from classifiers with confidence above a  threshold, or choosing label from the classifier prediction with highest confidence, but the aggregate strategy gave best results.  
The classifiers used are - Gaussian Process~\cite{GaussianProcess}, Naive Bayes~\cite{NaiveBaiyes}, SVM~\cite{SVM}, Decision Tree~\cite{DecisionTree}, Random Forest~\cite{RandomForest}, Multilayer Perceptron~\cite{MultiLayerPerceptron} and Adaboost ~\cite{Adaboost}.
We use scikit learn classifiers~\cite{scikit-learn} implementation with standard parameters.
}}

\subsubsection{Individual Fairness Test Case Generator}
We generate a set of synthetic samples using the synthetic test case generation procedure. For each sample, we change the predefined set of protected attribute (like race/gender) to create pair of test cases and which are checked against the model for label match. This closely resembles with the approach mentioned in \cite{THEMIS} which generates random samples (in comparison to our realistic samples) and subsequently perturbs them. 
\eat{
Once a discriminatory sample, $s$ is found, we generate further more test inputs in its neighborhood using the below perturbation function $p$.
We use an off-the-shelf explainer LIME~\cite{LIME} to get a set of top (say $50\%$) attributes, $X'$ such that $X'$$\subset$$X$, contributing to the explanation of test $s$. For every prominent attribute, say $x'$$\in$$X'$, we then generate further more local test inputs as per the below scheme based on its datatype.\\
\emph{Categorical}: Perturb $s.x'$ for all possible values of $x'$ to generate more tests. \\
\emph{Numerical}: Generate inputs by perturbing $s.x'$ for all integers or floats in range [$s.x'$- $\delta$, $s.x'$+$\delta$]. We have set $\delta$ as $3$ in our case. It is to be noted that the perturbation in such cases are bounded by the minimum and maximum permissible bounds of attribute as inferred from data constraints.
}

\subsubsection{Group Fairness Test Case Generator} 
Specifically just for the group fairness use case, we unconditionally make the \textit{protected attribute} independent by removing all those associations from the data constraints where it is mentioned as the target. This is done to ensure that the protected attribute in resultant synthetic test inputs follow the same frequency distribution as that of the input training data set. For example, if \texttt{gender} in the training data has composition of M:F in the ratio 2:1, then, it is desirable to test the model for group fairness with the test cases having same 2:1 M:F ratio. Note that this frequency distribution of protected attribute can be over-ridden by means of user-specified constraints as well as mentioned earlier in Subsection~\ref{sec:user-defined}.
\subsubsection{Robustness Test Case Generator}
For every realistic test sample $s$, we perform robustness testing of the input model by generating more inputs in its neighbourhood, and checking if the predictions of any neighbour is different than that of $s$. Here, we use the same definition of perturbation function $p$ as the one used in Individual Fairness Test Generator.





\section{Experimental Evaluation}
\label{sec:expt}
\noindent{\bf Benchmark \& Configuration.} We have assessed the performance of our approach on $10$ open-source data sets, namely, 
German~\cite{uci_repository},
Adult-8~\cite{adult-8},
Car~\cite{testing-fse},
US Exec~\cite{execution_data},
Iris~\cite{uci_repository}, Ecoli~\cite{uci_repository},
Cancer~\cite{uci_repository},
Instacart~\cite{codscomad-instacart}, 
Penbased~\cite{dataset}, and
Magic~\cite{dataset}.  Note that these datasets are used by prior state-of-the-art as well to report their evaluation. 
\cancel{These datasets are also used by an industrial AI platform to test their technology.}

 Our code is implemented as $\approx$1500 LOC of Python code.
All the experiments are performed in a machine running macOS 10.14, having 16GB RAM, 2.7Ghz CPU running Intel Core i7 running Python 3.7. For each benchmark, we have generated target model ($accuracy$$>$$85\%$) using default configuration as in \emph{scikit-learn}.\\

\noindent{\bf Experiment Goals.} By conducting our set of experiments, we broadly try to achieve three goals, as mentioned below.
\begin{itemize}[leftmargin=*]
\item{\textit{Realisticity Assessment -}}  
How realistic is the data generated using our synthesis algorithm on various metrics?



\item{\textit{Property Based Testing - }}
How well do the generated test cases contribute towards different property based testing?

\cancelpc{
\item{\textit{Coverage Based Testing -}} How well do the realistic tests perform while also considering path coverage constraints?
}
\end{itemize}

\noindent{\bf Realisticity Assessment.} This subsection discusses the results of various experiments performed to evaluate our realistic test generation approach. \eat{\cancel{
We use association-range anomaly, density anomaly, $JS$-divergence~\cite{jsdivergence} and model accuracy metrics to evaluate the realisticity of the synthetic data.}}
We compare the results on association-range anomaly, density anomaly and $JS$-divergence~\cite{jsdivergence} with CTGAN and TVAE~\cite{ctgan} synthesizers shown in Table~\ref{tab:real_assess}, and compare the results on model accuracy (a use-case of synthetic data other than testing) with the input training data shown in Table~\ref{tab:acc-realistic}. The metrics are described below.

\emph{Association-Range Anomaly:}
We generate synthetic data for $1000$ samples and
find the samples which are out of range 
with respect to the given data. This includes not only samples with out of range numerical column values but also samples with numerical values out of range for values in categorical columns. For example, 
for a sample with \texttt{education=primary}, the \texttt{salary} value must be in range for \texttt{education=primary}, which can be different from \texttt{salary} range for \texttt{education=tertiary}.

\emph{Density Anomaly:}
We generate synthetic data of size $1000$ and 
find anomalies
using a KNN classifier~\cite{PYOD-KNN-1} trained on the input data.
The classifier sets a threshold based on the euclidean distance of $k$th neighbour,
which is used as the outlying score. 
\emph{$JS$-Divergence Score:}
We generate $1000$ synthetic data samples and use $JS$-Divergence metric to find the difference in distributions for all columns in the given and synthetic data, averaged over all columns. We also find \emph{Association $JS$-Divergence} for subset of numerical columns filtered based on categorical column values. For example, for a  \texttt{salary} column, we find $JS$-divergence for \texttt{salary} values of  \texttt{education=primary} and \texttt{salary} values for  \texttt{education=tertiary}.
%
We duplicate either the input data or synthetic one, whichever is smaller, to make them of same size for comparison.

\emph{Model Accuracy:}
We divide the input data in $d_{train}$ and $d_{test}$ using the chosen split, and 
train a model $M_1$ using $d_{train}$ . We generate synthetic data $d_{\system{}}$ using the data constraints from $d_{train}$ having the same size as that of $d_{train}$. We generate the labels in $d_{\system{}}$ using $M_1$'s predictions and then, train another model $M_2$ using this $d_{\system{}}$. 
We compare the accuracy of models $M_1$ and $M_2$ on $d_{test}$. 

The results in Table~\ref{tab:real_assess} show that the \system{}-generated synthetic data has almost all of the samples for all $10$ datasets in range with respect to given data, as compared to CTGAN and TVAE, which have more than $10\%$ out of range samples for $7$ datasets, and more than $30\%$ out of range samples for $4$ datasets. The density anomaly results show that AITEST outperforms CTGAN and TVAE in $6$ out of $10$ datasets. 
The $JS$-Divergence results show that AITEST, similarly to CTGAN and TVAE, preserves the distribution of columns and
the associations. 
The average difference in $JS$-Divergence for AITEST and CTGAN/TVAE  over all datasets is $2.3\%$ for columns and $2.6\%$ for associations.
\eat{
Note that the $JS$-Divergence scores are below reasonable range and do not change even if the size of generated data is large. This is due to the fact that we preserve not only the correctness properties of the data using associations, but also the original distribution of the data, as described in Section~\ref{sec:Synth}.
Ideally this score should be $0$, but since there are multiple constraints to be satisfied, especially for columns that are highly correlated with many other columns, the score is greater than $0$. So, for datasets with large number of  correlated columns, 
the score is on a higher side. }

The model accuracy results in Table~\ref{tab:acc-realistic} show that the accuracy of model trained on synthetic data generated using \system{} is similar to the model trained on the given data for all datasets. 
Note that the average accuracy difference between the models trained on synthetic and input data  is less than $3\%$ for $24$ experiments out of $40$ ($10$ datasets $\times$ $2$ models $\times$ $2$ splits).



\begin{table}[tb]
\footnotesize
\caption{Comparative Realisticity Assessment}
\begin{center}
\begin{tabular}{|l||c|c|c|c|}
    \hline
    Bench. & \multicolumn{4}{c|}{\system{} vs CTGAN vs TVAE}\\ \cline{2-5}
     & Assoc.-Range& Density & JS-Div. & Assoc. JS-Div.\\
    \hline
    Iris
    &0, 970, 852
    &63, 798, 760
    &0.18, 0.18, 0.17
    &0.11, 0.15, 0.16\\
    
    Ecoli
    &0, 400, 453
    &10, 90, 5 
    &0.17, 0.13, 0.14 
    &0.13, 0.12, 0.14 \\

    Cancer
    &0, 935, 70
    &916, 237, 458
    &0.24, 0.19, 0.19
    &0.22, 0.20, 0.18\\ 

    Penbased
    &0, 918, 784
    &998, 651, 274
    &0.35, 0.32, 0.34
    &0.35, 0.28, 0.30\\ 

    Magic
    &0, 74, 16
    &992, 321, 944
    &0.25, 0.21, 0.19
    &0.27, 0.22, 0.24\\ 
    
    Adult-8
    &0, 26, 49 
    &9, 57, 30 
    &0.35, 0.35, 0.32 
    &0.22, 0.18, 0.18\\ 

    Car
    &0, 110, 557
    &1, 157, 607
    &0.49, 0.49, 0.57
    &0, 0, 0\\ 

    US Exec
    &6, 420, 102
    &8, 41, 1000
    &0.59, 0.57, 0.29
    &0.26, 0.32, 0.33\\ 

    German
    &0, 815, 331
    &6, 23, 2
    &0.34, 0.34, 0.31
    &0.22, 0.27, 0.21\\

    Instacart
    & 0, 113, 102
    & 563, 180, 284
    & 0.32, 0.31, 0.30
    & 0.33, 0.33, 0.31\\
    \hline
\end{tabular}
\end{center}
\label{tab:real_assess}
\vspace{-5mm}
\end{table}

\cancel{
\begin{table}[tb]
\footnotesize
\caption{Comparative Realisticity Assessment}
\begin{center}
\begin{tabular}{|l||r|l|l|l|}
    \hline
    Benchmark  & Metric & \system & CTGAN & TVAE\\ \hline
   
    \cancel{
    \multirow{4}{*}{Iris}
    &#Association-Range Anomaly&0&970&852\\ 
    &Density Anomaly&63&798&760\\ 
    &JS-Divergence&0.181&0.177&0.171\\ 
    &Association JS-Divergence&0.112&0.149&0.163\\ 
    \hline
    \multirow{4}{*}{Ecoli}
    &#Association-Range Anomaly&0&400&453\\ 
    &Density Anomaly&10&90&5\\ 
    &JS-Divergence&0.177&0.135&0.140\\ 
    &Association JS-Divergence&0.127&0.126&0.143\\ 
    \hline
    \multirow{4}{*}{BreastCancer}
    &#Association-Range Anomaly&0&935&70\\ 
    &Density Anomaly&916&237&458\\ 
    &JS-Divergence&0.241&0.197&0.187\\ 
    &Association JS-Divergence&0.217&0.206&0.178\\ 
    \hline
    \multirow{4}{*}{Penbased}
    &#Association-Range Anomaly&0&918&784\\ 
    &Density Anomaly&998&651&274\\ 
    &JS-Divergence&0.346&0.317&0.337\\ 
    &Association JS-Divergence&0.346&0.287&0.303\\ 
    \hline
    \multirow{4}{*}{Magic}
    &#Association-Range Anomaly&0&74&16\\ 
    &Density Anomaly&992&321&944\\ 
    &JS-Divergence&0.253&0.207&0.185\\ 
    &Association JS-Divergence&0.266&0.221&0.238\\ 
    \hline}
    \multirow{4}{*}{Adult-8}
    &#Association-Range Anomaly&0&26&49\\ 
    &Density Anomaly&9&57&30\\ 
    &JS-Divergence&0.349&0.349&0.323\\ 
    &Association JS-Divergence&0.224&0.179&0.173\\ 
    \hline
    \multirow{4}{*}{Car Rentals}
    &#Association-Range Anomaly&0&110&557\\
    &Density Anomaly&1&157&607\\ 
    &JS-Divergence&0.494&0.492&0.575\\ 
    &Association JS-Divergence&0&0&0\\ 
    \hline
    \multirow{4}{*}{US Executions}
    &#Association-Range Anomaly&6&420&102\\ 
    &Density Anomaly&8&41&1000\\ 
    &JS-Divergence&0.594&0.567&0.293\\ 
    &Association JS-Divergence&0.260&0.321&0.332\\ 
    \hline
    \multirow{4}{*}{German Credit}
    &#Association-Range Anomaly&0&815&331\\
    &Density Anomaly&6&23&2\\ 
    &JS-Divergence&0.339&0.334&0.306\\ 
    &Association JS-Divergence&0.222&0.270&0.205\\
    \hline
    \cancel{
    \multirow{4}{*}{Instacart}
    &#Association-Range Anomaly & 0 & 113 & 102\\
    &Density Anomaly & 563 & 180 & 284\\ 
    &JS-Divergence& 0.327 & 0.317 & 0.299 \\ 
    &Association JS-Divergence & 0.332 & 0.328 & 0.309\\
    \hline}
\end{tabular}
\end{center}
\label{tab:real_assess}
\vspace{-5mm}
\end{table}
}

\begin{table}[tb]
\footnotesize
\centering
\caption{Accuracy of Models trained with Synthetic Data}
\begin{tabular}{|l||c|c||c|c|}\hline
 Benchmark &  \multicolumn{4}{c|}{Train vs Synth for variable train-test split ratio}\\\cline{2-5}
  &  \multicolumn{2}{c||}{RF Accuracy (\%)} &  \multicolumn{2}{c|}{DT Accuracy (\%)}\\\cline{2-5}
  &      70:30&80:20 & 70:30&80:20 \\\hline
 Iris
    &91.11, 91.11
    &93.33, 93.33
    &96.66, 90.00
    &96.66, 96.66\\

  Ecoli
    &85.14, 84.15
    &81.18, 69.30
    &85.29, 85.29
    &79.41, 83.82\\

 Cancer
    &95.32, 88.30
    &91.81, 90.64
    &93.85, 93.85
    &90.35, 91.22\\

 Penbased 
    &98.57, 81.98
    &96.69, 86.11
    &98.72, 84.81
    &96.68, 82.40\\

Magic
    &87.57, 73.36
    &82.10, 69.90
    &86.96, 72.87
    &82.72, 68.24\\
    
Adult-8
    &80.25, 81.15
    &78.35, 76.71
    &79.89, 81.25 
    &78.41, 75.66\\

Car
    &84.25, 80.14
    &82.88, 79.45
    &85.71, 83.67
    &84.69, 82.65\\

German
    &77.33, 72.67
    &71.67, 61.67
    &74.00, 74.50
    &70.00, 69.00\\

US Exec
    &85.55, 84.70
    &82.44, 82.44
    &83.05, 83.05
    &82.20, 66.53\\

Instacart
    &56.46, 56.31
    &55.71, 52.57
    &56.39, 56.36
    &55.68, 52.77\\

  \hline
\end{tabular}
\newline
{$RF$-Random Forest, $DT$-Decision Tree}
\label{tab:acc-realistic}
\end{table}

\cancel{
\begin{table}[tb]
\footnotesize
\centering
\caption{Accuracy of Models trained with Synthetic Data}
\begin{tabular}{|l|l||r|r||r|r|}\hline
  Benchmark & Split & \multicolumn{2}{c||}{RF Accuracy (\%)} &  \multicolumn{2}{c|}{DT Accuracy (\%)}\\\cline{3-6}
  &         & Train&Synth & Train&Synth \\\hline
   \multirow{2}{*}{Iris}
    &70:30&91.11&91.11&93.33&93.33\\
    &80:20&96.66&90.00&96.66&96.66\\\cline{2-6}

\hline
   \multirow{2}{*}{Ecoli}
    &70:30&85.14&84.15&81.18&69.30\\
    &80:20&85.29&85.29&79.41&83.82\\\cline{2-6}

\hline
   \multirow{2}{*}{Breast Cancer}
    &70:30&95.32&88.30&91.81&90.64\\
    &80:20&93.85&93.85&90.35&91.22\\\cline{2-6}

\hline
   \multirow{2}{*}{Penbased}
    &70:30&98.57&81.98&96.69&86.11\\
    &80:20&98.72&84.81&96.68&82.40\\\cline{2-6}

\hline
   \multirow{2}{*}{Magic}
    &70:30&87.57&73.36&82.10&69.90\\
    &80:20&86.96&72.87&82.72&68.24\\\cline{2-6}
    
\hline
   \multirow{2}{*}{Adult-8}
    &70:30&80.25&81.15&78.35&76.71\\
    &80:20&79.89&81.25&78.41&75.66\\\cline{2-6}

\hline
    \multirow{2}{*}{Car Rentals}
    &70:30&84.25&80.14&82.88&79.45\\
    &80:20&85.71&83.67&84.69&82.65\\\cline{2-6}

\hline
    \multirow{2}{*}{German Credit}
    &70:30&77.33&72.67&71.67&61.67\\
    &80:20&74.00&74.50&70.00&69.00\\\cline{2-6}

\hline
    \multirow{2}{*}{US Executions}
    &70:30&85.55&84.70&82.44&82.44\\
    &80:20&83.05&83.05&82.20&66.53\\\cline{2-6}

\hline
    \multirow{2}{*}{Instacart}
    &70:30&56.46&56.31&55.71&52.57\\
    &80:20&56.39&56.36&55.68&52.77\\\cline{2-6}

  \hline
\end{tabular}
\newline
{$RF$-Random Forest, $DT$-Decision Tree}
\label{tab:acc-realistic}
\end{table}
}
\vspace{\baselineskip}
\noindent {\bf Property-based Testing.}
\begin{table}[tb]
\footnotesize
\centering
\caption{Property Testing: $Test$ vs $\system{}$ with DC only}
\begin{tabular}{|l|l||r|r||r|r||r|r|}\hline
  Benchmark & Split & \multicolumn{2}{c||}{RS (\%)} & 
  \multicolumn{2}{c||}{SS (\%)} & \multicolumn{2}{c|}{DI}\\\cline{3-8}
  &&Test& AIT & Test& AIT  & Test& AIT \\
    \hline
    \multirow{2}{*}{Adult-8: RF}
    &70:30&43.99&47.96&84.56&83.41&0.29&0.55\\
    &80:20&43.79&47.86&83.94&82.79&0.29&0.68\\\cline{1-8}
 
     \multirow{2}{*}{Adult-8: DT}
    &70:30&31.7&25.29&85.98&86.82&0.41&0.76\\
    &80:20&33.76&28.23&85.83&84.71&0.4&0.65\\\cline{1-8}
   
     \multirow{2}{*}{German: RF}
   &70:30&45.33&33.67&94.33&91.33&0.96&0.8\\
    &80:20&44.5&31&93&90.3&0.97&0.85\\\cline{1-8}
    
    \multirow{2}{*}{German: DT}
   &70:30&24&16.67&99&98&0.97&0.86\\
    &80:20&25.5&16&98&96.5&1.01&0.94\\\cline{1-8}
    
    \multirow{2}{*}{US Exec: RF}
   &70:30&72.52&88.95&96.03&99.43&1.05&1.02\\
    &80:20&79.24&88.56&95.34&97.88&1.08&1.02\\\cline{1-8}
    
    \multirow{2}{*}{US Exec: DT}
   &70:30&49.86&66.01&100&100&1.17&1.1\\
    &80:20&45.76&60.59&98.31&100&1.17&0.88\\\cline{1-8}
    
    \multirow{2}{*}{Car: RF}
   &70:30&7.53&4.35&71.23&71.21&1.76&1.99\\
    &80:20&2.04&1.79&72.45&70.69&0.76&1.93\\\cline{1-8}
   
    \multirow{2}{*}{Car: DT}
   &70:30&0.68&1.25&84.93&79.65&3.64&3.83\\
    &80:20&0&0&74.49&72.22&1.13&1.8\\\cline{1-8}
\hline
\end{tabular}
\newline
{AIT-$\system{}$, $RS$/$SS$-Robustness/Success Score, $DI$-Disparate Impact}
\label{tab:testvsdc}
\end{table}
In this subsection, we evaluate the effectiveness of our test generation approach to test an input AI model for different properties in following two modes of generation.
\begin{itemize}
    \item Synthesis using Data Constraints only (DC)
    \item Synthesis using Data Constraints along with User-Defined Constraints (DC + UDC)
\eat{\cancel{    \item Synthesis using Data Constraints along with Path Coverage constraints (DC + PC)}}
\end{itemize}

\paragraph{Synthesis using DC}
 In Table~\ref{tab:testvsdc}, we compare the effectiveness of \system{}-generated test data (using only the data constraints) with tests from train-test split for the varied set of properties. We expect that the synthesized test data will be equally effective or better for testing such properties. Next, we describe the relevant metrics to evaluate different test properties used in our set of experiments.
 
 \begin{itemize}[leftmargin=*]
     \item \emph{Adversarial Robustness} -
     We use \emph{Robustness Score} (i.e. $RS=\#Succ/\#Gen$) as an evaluation metric, where $Succ$ denotes the subset of the generated test cases ($Gen$) which fail robustness testing. 
      \item \emph{Individual Fairness} -
     We use \emph{Success Score} (i.e. $SS=\#Disc/\#Gen$)~\cite{testing-fse} as an appropriate metric to evaluate individual discrimination present in the model. Here, $Disc$ denotes the subset of the generated test cases ($Gen$) which results in individual discrimination. Higher the value of $SS$ and $RS$, better is the test set in uncovering faults for individual fairness and robustness, respectively.
     \item \emph{Group Discrimination} - We use \emph{Disparate Impact} (DI)~\cite{DI} which is a well known metric to evaluate group fairness. As per the industry standards, any test suite with $DI$$<$$0.8$ is treated as the one successful in uncovering group bias. 
\end{itemize}
 
 \eat{
 We divide the input data in $d_{train}$ and $d_{test}$ using the chosen split. We fetch the data constraints present in $d_{train}$ and synthesize realistic test inputs $d_{\system{}}$ with required sample count set as size of $d_{test}$. While synthesizing realistic data to test for group fairness property, we make sure that the protected attribute in the generated data has \cancel{the exact }similar distribution as in $d_{train}$. We, therefore, discard all the incoming associations to the protected attribute during the generation process. For individual fairness test case generation, we first synthesize a set of seed samples satisfying the inferred data constraints, out of which the discriminatory ones are then perturbed in its neighborhood to generate even more realistic test cases (refer Section~\ref{subsec:prop-test-synth}). The synthesized realistic inputs are then tested for different properties to report relevant metrics.}
 Note that we consider only one protected attribute at a time per benchmark. However, the effectiveness of \system{} will not be hampered even by considering multiple protected attributes. 
 
 \eat{
\cancel{ Table~\ref{tab:testvsdc} results show  that \system{}, when used in \emph{DC only} mode, performs equal or better in uncovering faults than the randomly picked 30\% or 20\% of sample tests from input benchmark for all the three properties. This is clearly in line with the expectations.}} We observe the following from Table~\ref{tab:testvsdc}:
 \begin{itemize}[leftmargin=*]
     \item  On an average across all benchmarks and models, robustness success rate offered by \system{} is $\approx$$35\%$, while it is $\approx$$34\%$ for tests from the chosen test split.
     \item \system{} on an average ($\approx$$88\%$) is equally effective at finding individual discrimination than the tests from test splits across all model variants for all benchmarks.
     \item For the models, such as related to Adult-8, showing group discrimination with DI for test-split less than $0.8$, \system{}-generated tests \cancel{also} yield a $DI$$<$$0.8$, and hence,\cancel{the latter is also capable of uncovering} uncover group discrimination. 
 \end{itemize}

 We conclude from these experiments that test data synthesis (without considering UDC \cancelpc{and path coverage}) is equally (or more) effective for discovering faults. This is attributed to the reason that our generation procedure slightly deviates from the actual distribution of the training data, and in most cases, such deviation helps in finding more faults. We believe that in real industrial scenarios, the real payload data may not follow the exact distribution present in the training data, but such test data synthesis can help to uncover more faults.

\begin{table}[tb]
\footnotesize
\centering
\caption{Property Testing: Test vs \system{} with DC+UDC}
\begin{tabular}{|l|l||r|r||r|r||r|r|}\hline
  Bench. & M:F & \multicolumn{2}{c||}{RS (\%)} &  \multicolumn{2}{c||}{SS (\%)} &  \multicolumn{2}{c|}{DI}\\\cline{3-8}
  &         & Test&AIT & Test&AIT & Test&AIT \\\hline

   \multirow{5}{*}{Adult-8}
    &1:1&66.84&78.33&14.31&21.22&0.411&0.847\\
    &2:1&66.84&76.43&14.31&16.50&0.411&0.772\\
    &1:2&66.84&74.04&14.31&29.03&0.411&0.778\\
    &1:3&66.84&79.24&14.31&23.72&0.411&0.848\\
    &3:1&66.84&78.48&14.31&14.92&0.411&0.905\\\cline{2-8}
    \hline
    \multirow{5}{*}{Car}
    &1:1&98.98&96.55&24.49&31.67&1.135&0.242\\
    &2:1&98.98&96.36&24.49&27.59&1.135&2.235\\
    &1:2&98.98&98.25&24.49&32.14&1.135&3.676\\
    &1:3&98.98&100&24.49&32.69&1.135&-1\\
    &3:1&98.98&98.04&24.49&32.08&1.135&4.421\\\cline{2-8}
    \hline
    
    \multirow{5}{*}{US Exec}
    &1:1&55.93&40.25&2.12&1.22&1.207&1\\
    &2:1&55.93&38.14&2.12&0&1.207&0.987\\
    &1:2&55.93&47.88&2.12&0&1.207&0.911\\
    &1:3&55.93&36.6&2.12&0.42&1.207&0.940\\
    &3:1&55.93&41.1&2.12&1.3&1.207&0.959\\\cline{2-8}
    \hline
    
    \multirow{5}{*}{German}
    &1:1&71.5&87&1&8&1.002&0.940\\
    &2:1&71.5&75&1&7&1.002&0.877\\
    &1:2&71.5&83.5&1&10&1.002&0.848\\
    &1:3&71.5&83&1&6.5&1.002&0.989\\
    &3:1&71.5&83.5&1&5&1.002&0.810\\\cline{2-8}
    \hline

\end{tabular}
\newline
\footnotesize{AITEST, All runs are for Decision Tree with 80:20 split}
\label{tab:udc}
\end{table}
\paragraph{Synthesis using DC+UDC}
In this experiment, we synthesize test inputs using data constraints along with our user-defined constraints where we override the Male-Female ratio (M:F) for gender (protected attribute) in data constraints by a different one. Note that here we intend to guage the deviation that synthesis using UDCs can bring, and therefore, for this experiment, we compare $d_{test}$ against the \system-generated samples satisfying the input data constraints along with the given UDC. Hence, for individual fairness testing, no neighborhood-based perturbation is carried out in this experiment.
The results for different metrics are recorded in Table~\ref{tab:udc} for different models trained on different benchmarks.
\begin{itemize}[leftmargin=*]
    \item With the variation in M:F ratios, an average improvement of $\approx$$5\%$ and $\approx$$2\%$ is recorded in Robustness and Success Score, resp., over the random test-split.
    \item $55\%$ models show an improvement of $>$$5\%$ in Individual Discrimination testing, with \texttt{Adult-8 M:F=1:2} showing the maximum gain of $\approx$$15\%$.
    \item $55\%$ models show a positive gain in Robustness testing with an average of $\approx$$10\%$.
    \item \system{}-generated tests for \texttt{Car} benchmark with UDC \texttt{M:F=1:1} uncovers group discrimination with $DI$ as $0.242$ ($<$$0.8$) which remains hidden while testing with original test-split ($DI$$=$$1.135$).
    \item Synthesis using UDC generates varied test suites which deviates relative to the  original test-split (i.e. $|AITEST-Test|/Test$) on an average of $\approx$$194\%$ for Individual Discrimination,  $\approx$$15\%$ for Robustness, and $\approx$$78.5\%$ for Group Discrimination testing. This significant deviation ascertain the need for UDC based testing to determine trustworthiness of the model under what-if scenarios.
    
\end{itemize}
\cancelpc{
\noindent{\bf Importance of Coverage-based Testing.} Here, we examine the benefit of  path coverage constraints during generation of test inputs. To perform this experiment, we learn a Random Forest or Decision Tree Classifier with a precision of 85\%-97\% for each benchmark. We split the input data in 80:20 to get $d_{train}$ and $d_{test}$, respectively. Using this trained model $M$ and $d_{train}$, we then learn a surrogate decision tree $S$ using TREPAN with $S_{fidelity}$$>$$90\%$. We use \emph{fidelity} as a metric to judge how well the surrogate mimics the input target model. It is defined as the fraction of the test inputs for which both the surrogate and input model output the same decision. Note that we have set the minimum sample count needed for split decision as $30$ in our TREPAN configuration while conducting all experiments. As per our evaluation on our benchmarks in Table~\ref{tab:trepan_perf}, our TREPAN implementation is efficient at inferring decision paths from the target model $M$ with an average fidelity of $\approx$$92\%$ across varied benchmarks for different models. Also, average prediction accuracy of the surrogate $S$ ($\approx$$81.9\%$) doesn't differ much from the target model $M$'s accuracy ($\approx$$82.3\%$).

\begin{table}[tb]
\scriptsize
\caption{Effectiveness of TREPAN}
\begin{center}
\begin{tabular}{|l||r|r|r|r|r|}
    \hline
    Benchmark  & Model M & $M_{acc}$  & $S_{acc}$ & $S_{fidelity}$ \\
    \hline
    \multirow{3}{*}{US Exec}
    &RF&86.02\%&86.02\%&100\%\\
    &LR&86.02\%&86.02\%&100\%\\
    &KNN&84.32\%&82.63\%&95.34\%\\
    \hline
    
    \multirow{3}{*}{German}
  &RF&72\%&72\%&99\%\\
    &LR&66.5\%&71.5\%&85\%\\
    &KNN&66.5\%&66\%&71.5\%\\
    \hline
    
    \multirow{3}{*}{Adult-8}
    &RF&76.23\%&76.23\%&100\%\\
    &LR&78.29\%&78.12\%&97.56\%\\
    &KNN&77.95\%&78.49\%&89.30\%\\
    \hline
    
    \multirow{3}{*}{Cancer}
    &RF&94.74\%&93.86\%&99.12\%\\
    &LR&96.49\%&93.86\%&95.61\%\\
    &KNN&91.23\%&91.23\%&98.25\%\\
    \hline
    
    \multirow{3}{*}{Car}
    &RF&65.31\%&65.31\%&100\%\\
    &LR&64.29\%&64.29\%&100\%\\
    &KNN&71.43\%&71.43\%&97.96\%\\
    \hline
    
    \multirow{3}{*}{Iris}
    &RF&90\%&90\%&100\%\\
    &LR&90\%&96.67\%&93.33\%\\
    &KNN&93.33\%&96.67\%&96.67\%\\
   
    \hline
    
    \multirow{3}{*}{Ecoli}
    &RF&80.88\%&76.47\%&92.65\%\\
    &LR&85.29\%&76.47\%&88.24\%\\
    &KNN&86.76\%&86.76\%&88.24\%\\
       
    \hline
     \multirow{3}{*}{Penbased}
     &RF&74.03\%&73.62\%&96.95\%\\
    &LR&92.68\%&90.68\%&92.59\%\\
    &KNN&99.5\%&96.09\%&95.86\%\\
   
    \hline
     \multirow{3}{*}{Magic}
     &RF&73.79\%&73.5\%&99.5\%\\
    &LR&78.68\%&77.92\%&96.82\%\\
    &KNN&78.81\%&80.44\%&83.96\%\\
   
    \hline
     \multirow{3}{*}{Bank Market}
     &RF&88.43\%&88.43\%&100\%\\
    &LR&89.45\%&89.15\%&98.06\%\\
    &KNN&89.59\%&89.30\%&91.05\%\\
   
\hline
    
\end{tabular}

\footnotesize{$RF$-Random Forest, $LR$-Logistic Regression, $KNN$-k-Nearest Neighbors}
\end{center}
\label{tab:trepan_perf}
\vspace{-5mm}
\end{table}

We record the total count of unique decision paths present in this surrogate $S$ as $Total$ and the count of unique decision paths traversed by all the test inputs in $d_{test}$ in $Test$ column. Using the path coverage constraints inferred from $S$ and data constraints inferred from training inputs $d_{train}$, we then generate exactly same number of synthetic test inputs (Algorithm DC+PC) as there were in original test set $d_{test}$ using \system{}. We then record the count of unique decision paths traversed by these synthetic test inputs as $AIT$.
Apart from this, we also record the count of unique paths traversed by test cases from both $d_{test}$ and the \system{}-generated ones which \emph{fail} individual discrimination and robustness testing. We summarize the  key observations from  Table~\ref{tab:cov_imp} below:
\begin{itemize}[leftmargin=*]
    \item  On an average across all models \eat{trained on different benchmarks}, the test inputs generated using \system{} offers $\approx$$51\%$ more path coverage than test-split ones. A significant improvement in path coverage $\approx$$[52\%-64\%]$ is recorded for different models trained with German, Adult-8 and US Exec benchmarks.
    \item \system{}-generated tests fail individual discrimination and robustness testing on an average of $\approx$$16\%$ and $\approx$$45\%$, resp., more decision paths than the test-split.
\end{itemize}
Thus, we can conclude that \system{} performs much better at generating diverse tests covering varied decision paths than the random ones. This stands valid for both the overall set of generated tests, and their failure subsets for both individual discrimination and robustness testing as well.

\begin{table}[tb]
\footnotesize
\centering
\caption{Path Coverage: Test vs \system{} with DC+PC}
\begin{tabular}{|l|l|l||r|r||r|r||r|r|}\hline
  Bench. & M &  $Total$ & \multicolumn{2}{c||}{All Tests} &  \multicolumn{2}{c||}{IndDisc Fails} & \multicolumn{2}{c|}{Robust Fails} \\\cline{4-9}
  &      &   & Test&AIT & Test&AIT& Test&AIT \\\hline
  
  \multirow{2}{*}{Car}
    &RF&42&28&42&21&32&28&42\\
    &DT&46&31&46&20&31&31&46\\
 \hline
  \multirow{2}{*}{US Exec}
    &RF&161&58&161&5&18&26&102\\
    &DT&179&65&179&3&14&57&167\\
\hline
\multirow{2}{*}{German}
&RF&149&72&149&15&24&54&99\\
&DT&149&71&149&3&18&64&132\\
\hline
\multirow{2}{*}{Adult-8}
&RF&4171&1829&4171&702&1809&1596&3624\\
&DT&4409&1841&4409&555&1285&1739&4172\\
   \hline

\end{tabular}
\label{tab:cov_imp}
\vspace{-2mm}
\end{table}
}


\textit{Timings.} Constraints inference takes from a few seconds for a small input data to about 250 seconds for a dataset with $1$ million rows and $70$ columns.
The synthesizer takes from a few seconds for $100$ rows to  $\approx$$30$ seconds for $10,000$ rows.

\cancel{
\subsection*{Threats to Validity}

\begin{itemize}[leftmargin=*]
\cancel{    \item \emph{Relevance of benchmarks.} The benchmarks used are well-known in the testing and fairness domain, and used in related works too. \cancel{and in IBM product testing.}}
    \item \emph{Protected attributes.} For fairness testing experiments, we consider only one protected attribute at a time per benchmark. However, the effectiveness of \system{} will not be hampered even by considering multiple protected attributes. 
    \eat{ \cancel{and may even lead to reduced processing time if associations are neglected for protected attributes during group fairness test synthesis.}}
    \item \emph{Neighborhood function.} We use a local explainer, LIME to define the neighborhood function for robustness and fairness testing. LIME can handle data sets having numeric and category attributes, but not high dimensional data like image, sound or video. But, any sophisticated explainer such as Grad-CAM~\cite{gradcam} can come to rescue.
    \item \emph{Model Configuration.} The various models used in this paper is tuned by hand. The results may slightly vary for different hyper-parameter configurations.
\end{itemize}
}
\section{Related Work}
\label{sec:related}

This section discusses the existing works spread across two related spheres - model coverage and realistic data synthesis. 
\paragraph{Model Coverage.} Neuron coverage \add{(DeepXplore)} ~\cite{DEEPXPLORE} measures the percentage of neurons in a deep neural network that are activated. $K$-multisection neuron coverage (KMNC) and 
Strong Neuron Activation Coverage (SNAC) ~\cite{DEEPGAUGE} extend the idea of neuron coverage, measuring how thoroughly the given set of test inputs cover the range for a neuron and capturing the percentage of corner case regions that are covered by the set of test inputs, respectively. 
Neuron Boundary Coverage (NBC) and Top-$k$ Neuron Coverage (TKNC) ~\cite{DEEPGAUGE} are similar coverage criteria specific for deep neural networks measuring the coverage of corner cases by test data, and the fraction of top-$k$ neurons within a layer for given test data, respectively.
This is done by partitioning the region into sections between the boundaries, and measuring if each partition has been visited. 
It is claimed that good quality test datasets have neuron activation values spread across the boundaries and close to the corner regions. DeepCT~\cite{DEEPCT} considers the interactions of neurons and proposes a set of combinatorial testing criteria for DNNs.
Adequacy~\cite{ADEQUACY}, as a measure, studies the effects of features from the adjacent layer. Their intent comes from the fact that a deeper neural layer captures complex features and therefore, its next layer can be considered as its summary. These testing criteria mainly focus on feed-forward neural networks, while DeepStellar~\cite{DEEPSTELLAR} proposed the model-based  testing criteria for recurrent neural networks. 
 
All these coverage criteria are non-generic and specific to model architecture. We define a model agnostic coverage criteria which is universal and can be used for any model with a black-box access.

\paragraph{Realistic Data Synthesis}
Note that there exist sophisticated techniques, such as GAN~\cite{GAN} and VAE~\cite{VAE}, which can generate realistic synthetic data. But, such approaches suffer from the following problems which prohibit their usage in this practical setting. 1) GAN/VAE based approaches require to train a model which in-turn requires hyper-parameter tuning which is currently done by hand. Although there are \cancel{industry standard} approaches for auto-building classification models~\cite{AUTOAI}, but such  techniques are still not available for generative models. Compared to our approach, all these techniques are inflexible and cannot generate custom data for specific domains based on user-defined constraints. 
2) The inherent data constraints are not easily customizable - based on user-defined constraints \cancelpc{and path constraints} which are applicable for test data generation.  3) The inherent data constraints captured by such frameworks are not interpretable - so, if user wants to see the current distribution and customize it for test data generation, then it's not possible. We, therefore, use a two step technique for data-synthesis. In the first step, we capture the inherent feature/column and association constraints present in the data. The set of constraints are represented in form of a directed graph. \eat{\cancel{This is similar in spirit with Bayesian Networks~\cite{BN} involving categorical and numerical attributes.}} The graph can then be changed considering additional constraints, like user-defined constraints \cancelpc{and path constraints}. Finally, a sampling algorithm is employed to generate data from the graph. An another technique called SMOTE~\cite{SMOTE} is also prominently used to generate synthetic data, but it can generate only in the neighbourhood of the existing data. 

\section{Conclusion}
\label{sec:conc}

\cancel{\noindent {\bf Summary.}} We have presented a framework for black-box testing of AI models which involves 1) generation of realistic synthetic data with 2) model-agnostic path coverage, 3) user-controllable data  generation, and 3) enabling testing of fairness and robustness properties. 

Our main learning is that
state-of-the-art synthesis techniques are not always suitable for handling practical requirements such as  user-defined-constraints \cancelpc{and coverage}. Handling such scenarios increases in model trustworthiness under what-if scenarios and increased diversity in test results. 

\noindent {\bf Future Work.} We plan to investigate automated testing of unsupervised models and for modalities like text and image classifiers, and time-series predictive models.

\newpage
\bibliographystyle{ACM-Reference-Format}
\bibliography{IEEEabrv,bib}

\end{document}